\documentclass[runningheads]{llncs}
\usepackage[T1]{fontenc}
\usepackage{comment}

\usepackage{graphicx}
\usepackage{latexsym}
\usepackage{amssymb}
\usepackage{amsmath}
  \usepackage{amsthm}
\usepackage{booktabs}
\usepackage{enumitem}
\usepackage{graphicx}
\usepackage{color}
\usepackage{algorithm}
\usepackage{algpseudocode}

\usepackage{caption}
\usepackage{subcaption}

\algnewcommand\algorithmicforeach{\textbf{for each}}
\algdef{S}[FOR]{ForEach}[1]{\algorithmicforeach\ #1\ \algorithmicdo}

\newcommand{\BibTeX}{B\kern-.05em{\sc i\kern-.025em b}\kern-.08em\TeX}

\AtBeginDocument{%
  \providecommand\BibTeX{{%
    Bib\TeX}}}

\begin{document}

\begin{frontmatter}


\title{Pareto Optimal Algorithmic Recourse in Multi-cost Function}


\author{Wen-Ling Chen \and Hong-Chang Huang \and
Kai-Hung Lin \and Shang-Wei Hwang \and Hao-Tsung Yang}

\authorrunning{W. Chen et al.}
%
\institute{National Central University, Taiwan}
\maketitle              
%
\begin{abstract}

In decision-making systems, algorithmic recourse aims to identify minimal-cost actions to alter an individual's features, thereby obtaining a desired outcome. This empowers individuals to understand, question, or alter decisions that negatively affect them. However, due to the variety and sensitivity of system environments and individual personalities, quantifying the cost of a single function is nearly impossible while considering multiple criteria situations. Most current recourse mechanisms use gradient-based methods that assume cost functions are differentiable, often not applicable in real-world scenarios, resulting in sub-optimal solutions that compromise various criteria. These solutions are typically intractable and lack rigorous theoretical foundations, raising concerns regarding interpretability, reliability, and transparency from the explainable AI (XAI) perspective. To address these issues, this work proposes an algorithmic recourse framework that handles non-differentiable and discrete multi-cost functions. By formulating recourse as a multi-objective optimization problem and assigning weights to different criteria based on their importance, our method identifies Pareto optimal recourse recommendations. To demonstrate scalability, we incorporate the concept of $\epsilon$-net, proving the ability to find approximated Pareto optimal actions. Experiments show the trade-off between different criteria and the method's scalability in large graphs. Compared to current heuristic practices, our approach provides a stronger theoretical foundation and better aligns recourse suggestions with real-world requirements.


\end{abstract}

\end{frontmatter}




\section{Introduction}
\begin{figure}[ht]
\centering
\includegraphics[width=90mm]{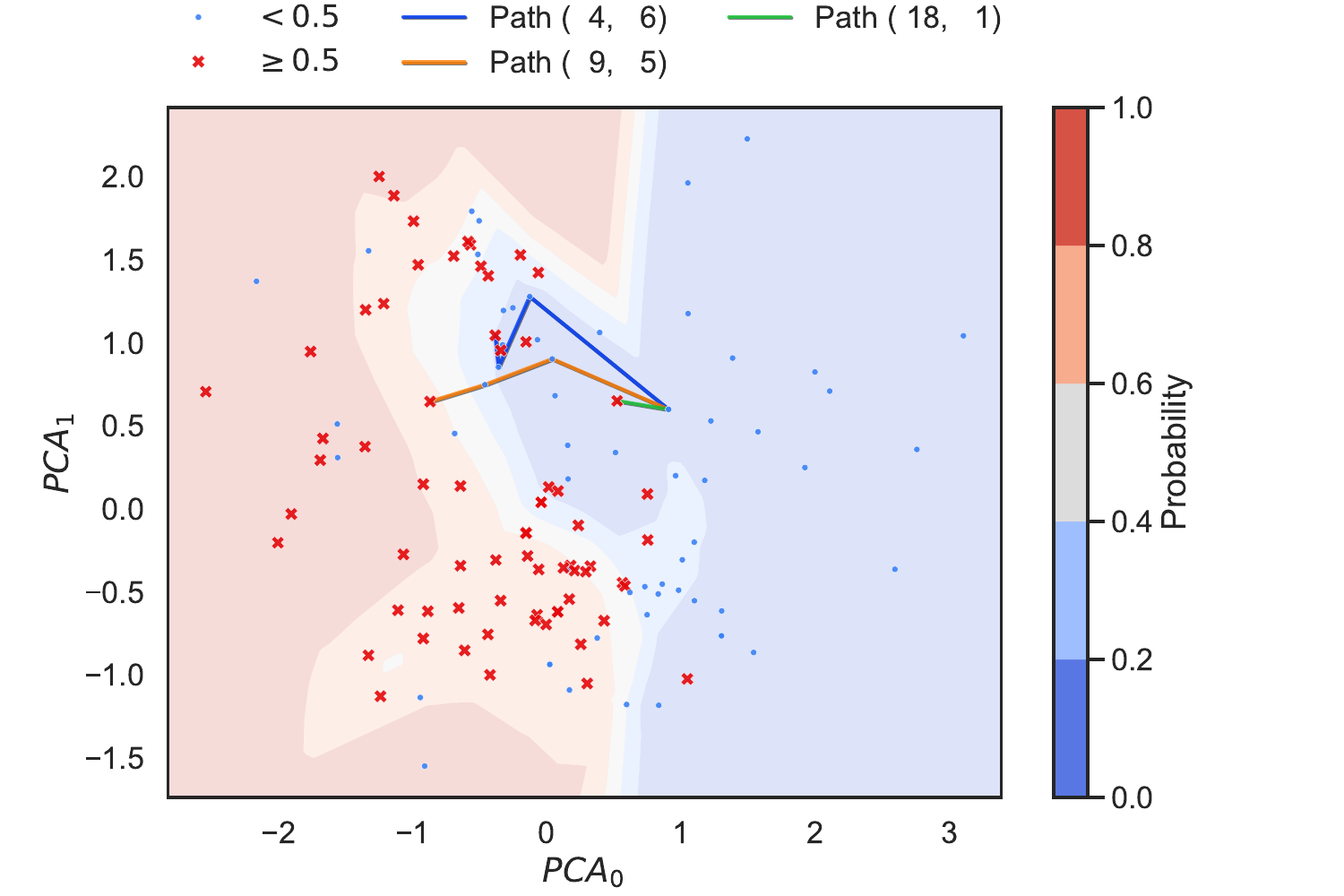}
\caption{In this example, our approach provides three different paths which are all pareto optimal for the recourse plans. Each path contains different criteria and the user can choose one that fits him the most.}
\label{fig:toy eg}
\vspace{0.2cm}
\end{figure}


Deep learning has been the most popular technique in Artificial Intelligence (AI)~\cite{pouyanfar2018survey},
~\cite{zhao2019object}, 
~\cite{wang2021deep}, 
~\cite{otter2020survey}, 
~\cite{danilevsky2020survey}, 
~\cite{turay2022toward}, ~\cite{tang2022perception}, and people increasingly rely on AI for generating decisions ~\cite{zhang2021artificial},~\cite{peres2020industrial}. However, the opacity of deep learning models complicates tracing and understanding decision rationales ~\cite{daston2021objectivity},~\cite{balasubramaniam2022transparency},~\cite{srivastava2022xai}. This issue becomes critical in systems involving safety or human activity, such as the 2018 Uber autonomous vehicle accident ~\cite{lawless2022toward}, racial bias in image restoration ~\cite{reconstructpictureofBarackObama}, or inappropriate advice from medical chatbots ~\cite{chatbottoldafakepatienttokillthemselves}. These incidents raise fundamental trust concerns in AI systems.

The use of recourse in decision-making and recommendation systems is driven by the need for transparency and accountability, allowing individuals to understand, question, or alter decisions that negatively impact them, such as in loan approvals, medical diagnostics, or education. By integrating recourse, AI systems align more with user needs, enabling participation and influence in decision-making, thereby promoting trust and fairness in AI applications ~\cite{chen2020simple}. Current algorithmic recourse methods are often gradient-based, utilizing computational efficiency to navigate high-dimensional solution spaces and identify necessary modifications ~\cite{fokkema2023attribution},~\cite{karimi2021algorithmic},~\cite{NEURIPS2020_8ee7730e},~\cite{upadhyay2021towards}. These methods incorporate constraints to ensure feasible and applicable changes, adhering to ethical and legal standards ~\cite{gardner2022responsibility},~\cite{venkatasubramanian2020philosophical}.

However, the application of recourse faces significant challenges~\cite{karimi2020survey}, particularly in navigating the high-dimensional space of user features to pinpoint the optimal recourse. In real-world scenarios, determining the most effective recourse is complicated by the multifaceted nature of user attributes and the limitations of existing methods, which often yield suboptimal solutions that compromise various goals~\cite{ustun2019actionable}. These methods lack thorough theoretical analysis and often rely on heuristic methods such as gradient descent, raising concerns about their interpretability and the transparency they are meant to enhance. In the following, we address those challenges in three directions.

\begin{enumerate}
\item \textbf{Multi-Cost Scenarios}: Dealing with multi-cost scenarios in algorithmic recourse methods is complex. Real-world decisions are influenced by multiple types of costs, such as money, time, and effort. Current methods often optimize a single cost, neglecting trade-offs between different costs. Gradient-based approaches handle bi-criteria scenarios by merging loss functions, but deriving suitable weights for multiple costs is unrealistic. Multi-cost scenarios require a nuanced approach to capture the interplay between different costs and provide balanced recommendations.

\item \textbf{Intractable recourse}. As mentioned above, many algorithmic recourse methods are gradient-based or heuristic-based, which lack robust theoretical underpinnings~\cite{karimi2020algorithmic}. This can lead to solutions that are not only suboptimal but also unreliable and unpredictable in their effectiveness. Recent work from Slack et al.~\cite{slack2021counterfactual} emphasizes the vulnerabilities of those recourse applications that are highly sensitive to small changes in the input and can be manipulated under a slight perturbation from an adversary. Hence, providing a solid theoretical framework and tractable solution is needed to assess the reliability, increase the interpretability, and directly influence the trustworthiness of the recourse.

\item \textbf{Simplistic Cost Functions}: Another challenge lies in the overly simplistic and unrealistic cost functions employed by many recourse methods~\cite{chereda2021explaining,dwork2012fairness,verma2020counterfactual}. These functions often assume smoothness and continuity, which may not accurately reflect the discrete and non-differentiable nature of real-world cost functions. For instance, in a loan application scenario, the cost of improving one's credit score cannot be modeled as a continuous function due to the discrete steps involved in credit score changes. The oversimplification of cost functions can lead to recommendations that are impractical or unattainable for individuals, thereby diminishing the utility of the recourse.
\end{enumerate}

\paragraph{Our contribution:} 
In this work, we propose a novel approach to algorithmic recourse to navigate the challenges of multi-cost scenarios and can be generalized to non-differentiable or discrete cost functions. Our method first constructs the actionability graph where each edge represents a possible action and the multi-cost is recorded on the edges. Then, we modified the shortest path algorithm to fit the multi-cost function and found all Pareto optimal paths, where each path represents a feasible action list under certain multi-criteria for the user. The analysis shows that our algorithm is optimal and the time complexity is polynomial on the graph size. We then provide a highly non-trivial method to scale our approach into a large graph via a shrinking procedure, which preserves the approximation factor of the Pareto optimal paths and utilizes $\epsilon$-net, which covers most of the points among all the shrunk graphs.

A toy example in Figure ~\ref{fig:toy eg} illustrates a model predicting whether a person's income exceeds $50K/yr$. For a specific user wanting to increase their income, our algorithm provides three Pareto optimal paths under two cost functions: years of action and change in education levels. Each path represents feasible actions balancing these costs.

There are a few works that use a path-oriented approach for the recourse action~\cite{poyiadzi2020face,nguyen2023feasible} but only focus on specific criteria such as diversity or the density of the path. Other work that uses mixed-integer programming to deal with multi-criteria~\cite{russell2019efficient,mothilal2020explaining} but only allows a limited number of cost functions to fit in. Dandl et al. propose a model-agnostic method that utilizes multi-objective evolutionary algorithm~\cite{dandl2020multi} which is close to our objective but again focuses on specific criteria only. To the best of our knowledge, we are the first work to discuss the multi-cost recourse plan where the cost functions are free to choose for the users as long as they are metric functions. Additionally, we propose a solution to the scalability problem which has not been discussed in any other path-oriented recourse approaches. Lastly, our proposed solution is simple, tractable, and interpretable bringing the trustworthiness of the system, which we emphasize as an important feature in the XAI community. Our experiments show that our algorithm can find different paths under different criteria and is scalable for large graphs.

\section{PROBLEM DEFINITION}
We consider a fixed predictive model $h: \mathcal{X} \rightarrow Y$, with $\mathcal{X}= \mathcal{X}^1 \times \mathcal{X}^2 \cdots \mathcal{X}^d$. Each attribute $\mathcal{X}_i$ can be either a continuous or discrete value (e.g., categories). The output can be either binary class $\mathcal{Y}=\{0,1\}$ or stochastic $\hat{\mathcal{Y}}=[0,1]$, which is the probability of the user classified into $1$. The 0 and 1 represent the negative outcome and positive outcome, respectively. A corresponding application example can be a loan approval system where 0 means the ``loan denied'' and 1 means ``loan approved.''

Given a set of $n$ accessible data points $\{x_1,x_2, \cdots x_n \}, x_i \in \mathcal{X}$ and a set of $k$ cost functions $C=\{c_1,c_2, \cdots c_k\}$, one can construct a directed graph which we called actionability graph, where each node is an accessible data point (such as training samples). There is a directed edge from $u$ to $v$ if a feasible action exists from $u$ to $v$. Each cost function $c_i: \mathcal{X} \times \mathcal{X} \rightarrow \mathcal{R}^+$ is a metric that represents the ``distance'' of two data points. Notice that, the cost functions can be non-differentiable, an attribute of the education level is discrete and non-differentiable. Now, consider a specific point $s$ where $h(s)=0$, a feasible path $P$ in the actionability graph is the one that every edge in this path is a feasible action. By abusing the notation of cost function $c$, we denote the total cost of path $P$ is $c(P)$ for the function $c$, which can be other aggregation functions as long as they are metrics, such as summation or any maximum function. 

We define path $P$ as dominated by path $Q$ if and only if for any cost function $c_i$, $c_i(P) \geq c_i(Q)$. Our objective is to find a set of feasible paths in the graph such that the end of each path is a recourse point $t$, where $h(t)=1$ and each path is Pareto optimal concerning all the cost functions. That is, taking summation as the example, path $P$ is Pareto optimal if and only if there does not exist a path $Q$ such that for any cost function $c$, $\sum_{(x,y) \in P} c(x,y) > \sum_{(x,y) \in Q} c(x,y)$.

\section{MULTI-CRITERIA RECOURSE}
\label{subsec:algo}
Our approach involves handling the minimization of cost for a single edge, which can be analogized to the concept of finding the shortest path. We manage this by maintaining and updating a table to capture the trade-offs among various types of costs. This table is instrumental in recording and managing the interplay of different cost criteria on different paths. In the following, we assume the model $h$ is deterministic but the algorithms and all the analysis can be applied to the stochastic model too.

Users are allowed to decide on the properties of the target instance - counterfactual. Under what circumstances is it considered a successful flip? What would be the criteria for a feasible path? The customized cost functions can be very flexible as long as they are metric functions. After that, the edges of the actionability graph are constructed by iterating through each pair of vertices in the input set $V$, checking if there is a possible action between them, and calculating the multi-cost. Unlike simple graphs, each constructed edge has several costs.

\subsubsection{Algorithm} we use the dynamic programming and modify Bellman-Ford algorithm~\cite{bellman1958routing} to find all the pareto optimal paths. The main goal is to handle the minimum cost within all considered criteria. A hyperparameter $\eta, 1 \leq \eta \leq n-1$ is introduced here to decide the maximum number of edges in a feasible path (i.e., the length of the path in terms of the number of hops). In reality, a path with too many edges may be redundant and hard to interpret. 

\begin{algorithm}
\caption{Pareto-shortest-path}
\label{algo:pareto-shortest}
\begin{algorithmic}[1]
    \Function{Pareto-paths}{$G = \{E,V\}, s$}
        \ForEach {$v \in V$}
            \State $D^0_v \gets \phi$ \Comment{Initialization}
        \EndFor
        \State$D^0_{s} \gets \{ <0,0,\cdots 0> \}$     
          
        \For{$\ell \gets 1$ to $\eta$}
            \ForEach{$(u,v) \in E$}
                \State update($D^\ell_v, (D^{\ell-1}_u, W_{uv})$) 
            \EndFor
        \EndFor
        \State \Return $\{D^\eta_t| t\in V \text{ and } h(t)=1\}$
    \EndFunction
\end{algorithmic}
\end{algorithm}

Algorithm~\ref{algo:pareto-shortest} computes the shortest paths from a given source vertex to all other vertices in a graph. $D^\ell_v$ denotes the Pareto table of $v$ that represents the set of current Pareto paths in iteration $\ell$ for reaching vertex $v$ from the source vertex. Line 2 to Line 5 are initialize steps. All items in $D^0_v$ are empty except for the source vertex. We set the distance to itself $D^0_s$ as zero for all cost dimensions. Line 6 to Line 11 uses Dynamic Programming to iteratively find all the Pareto paths. It iterates over the vertices excluding the source and conducts a tailored merging process, for each edge $(u, v)$ in the graph, performs the update operation. The update operation extends $D^{\ell-1}_v$ to $D^{\ell}_v$ by first concatenating the distance from the vertex $u$ with the multi-cost $W_{uv}$ and then pruning the dominated path. The concatenating step is the same as the standard operation in Bellman-Ford except each path in $D^{\ell-1}_u$ has multi-cost. The operation time here increased by at most $k$ times. The prune operation is employed to refine the multi-cost estimate of paths in $D^\ell_v$ by comparing the paths between $D_v^{\ell-1}$ and $D_u^{\ell-1}+W_{uv}$ and only record the non-dominated paths into $D_v^{\ell}$. Finally, Line 11 reports all the Pareto paths for all possible recourse points $t$, where $h(t)=1$.

\subsubsection{Analysis} In Algorithm~\ref{algo:pareto-shortest}, the time complexity analysis involves two main steps: concatenate and prune, which are repeated for at most $\eta$ iterations. Denote $\tau$ as the maximum size among all Pareto tables and $\gamma$ as the maximum degree among all vertices. In the concatenation step, for each vertex $v$, all the incoming neighbors $u$ of $v$ have been concatenated via $w_{uv}$, which takes at most $O(\gamma \tau)$. In the pruning step, $D_v$ is updated to keep the Pareto paths only. This is related to finding the maxima of a point set problem~\cite{preparata2012computational}. Naively, we can check all pairs of paths with each criterion to remove the dominated ones, which takes $O(k\gamma^2 \tau^2)$. However, since we are not taking an arbitrary point set but merging Pareto tables from all the incoming neighbors. During the pruning steps, the Pareto table is maintained via lexicographic orders. Thus, the running time can speed up, depending on the number of the cost functions $k$. When $k=2$, since the first cost is always ordered we can compare the second cost directly, which takes $O(\gamma \tau)$. When $k=3$, one can use basic sorting to remove dominated paths, which takes $O(\gamma \tau \log \gamma \tau)$. This can be further sped up by using the data structure of van Emde Boas tree~\cite{karlsson1988scanline}, takes $O(\gamma \tau \log \log \gamma \tau)$. When $k>3$, one can sort the rest of the criteria with lexicographic order, which increases the time complexity to $O(\gamma \tau \log^{k-4}\log\log \gamma \tau)$~\cite{gabow1984scaling}. Overall, the time complexity of Algorithm~\ref{algo:pareto-shortest} is $O(\gamma \tau \log^{k-4}\log\log \gamma \tau * \eta |E|)$, where $|E|$ represents the number of the edges. 

In reality, the size of the Pareto table $\tau$ is bounded by the number of different paths in the graph, this may vary depending on the specific problem instance and the number of criteria considered. Thus, one may face the scalability issue when the graph is large with high degrees. We address this issue in Section~\ref{sec: scalability} and provide a solution under those circumstances, the correctness of Algorithm~\ref{algo:pareto-shortest} is provided in the full version of this work. 

\begin{theorem}
    Algorithm~\ref{algo:pareto-shortest} finds all the Pareto optimal paths from the source to any endpoint $t$, where $h(t)=1$.
\end{theorem}

\section{SCALABILITY ENHANCEMENT}
\label{sec:scalability}
In Section~\ref{subsec:algo}, we show that finding all Pareto optimal on the actionability graph is crucial to the size of the Pareto tables $\tau$, edge size $|E|$, and number of iterations $\eta$. Since $\eta$ corresponds to the number of hops of the output paths, which is usually as a constant (considering in reality that a long feasible path is redundant and non-interpretable), the bottleneck of the running time is mainly on $\tau$ and $|E|$. Additionally, $\tau$ is bounded by the number of different paths from each pair of nodes, which highly depends on the size of the vertices and the connectivity. To decrease the graph size and simplify the connectivity structure, one idea is to shrink the vertices of the graph such that there are only a small number of ``representative'' nodes, and the shortest path in this shrunk graph still preserves or approximates the distance of the original graph. This is the idea of core-set from the computational geometry perspective (see survey in~\cite{agarwal2005geometric}). The challenge here is that the cost functions are not specific but highly general. Additionally,  we also need to incorporate all the $k$ cost functions to get the Pareto optimal. Fortunately, the computation of the cost functions usually contains some structures rather than arbitrary values for any pair of points. This inspires us to utilize the idea of $\epsilon$-net~\cite{haussler1986epsilon} to shrink the size of the graph and also ensure the quality. 

To explicitly explain our idea, we first define the notation of shrinkable.''

\begin{definition}
\label{def: shrinkable}
    Given $G=(V,E)$ and a cost function $c$, we say a vertex $i$ is $\kappa$-shrinkable to vertex $j$ if and only if $\forall (p,i) \in E$
    $$ 
    (p,j) \in E \text{ and } c(p,j) \leq \kappa c(p,i)
    $$
\end{definition}

Given the approximation factor $\kappa$, one can iteratively shrink all the shrinkable vertices in the graph until there is no shrinkable vertex anymore. We call this induced subgraph a shrunk graph $G_S$ and the one with the smallest cardinality is $G_S^*$. Obviously, any $G_S$  preserves $\kappa$-approximation factor. The shortest path between any pair of the nodes in $G$ has another path in $G_S$ which is at most $\kappa l$ times, where $l$ is the number of the hops of the path. However, finding the $G_S^*$ is highly non-trivial. It depends on the order of vertices in the shrinking procedure. A toy example is in the following. Consider a graph with vertices $\{p,i,j,r\}$ and edges $\{(p,i),(p,j),(p,r)\}$ where $c(p,i)=\kappa c(p,j)= \kappa^2 c(p,r)$, if $j$ shrinks to $i$, then $i,r$ are not shrinkable. On the other hand, if $i$ shrinks to $j$, then $r$ can shrink to $j$ too. Thus, we want to have another subgraph that catches most properties of $G_S$ and can be generated efficiently. This is the place where the $\epsilon$-net joins into our work. In the following, we will first introduce the formal definition of $\epsilon$-net and then show how to utilize it under our context.

\begin{definition}
\label{def: epsilon-net}
Given a range space $(\mathcal{X}$,$\mathcal{R})$, let $\mathcal{A} \subset \mathcal{X}$ be a finite subset, and $0<\epsilon<1$. Then a subset $\mathcal{N} \subset \mathcal{A}$ is called an $\epsilon$-net of $\mathcal{A}$ w.r.t to $\mathcal{R}$ if 

$$
\forall r \in \mathcal{R}, |r \cap \mathcal{A}|>\epsilon |\mathcal{A}| \rightarrow r\cap \mathcal{N} \neq \emptyset
$$
\end{definition}

Now, we define the $\epsilon$-net under our context. 

\begin{definition}
\label{def: our epsilon-net}
We say $G_\epsilon$ is an $\epsilon$-net of $G$ if for any vertex $v$ in some shrunk graph $G_S$ that is shrunk by more than $\epsilon n$ vertices, then either $v \in G_\epsilon$ or $u \in G_\epsilon$, where $v$ is shrinkable to $u$.    
\end{definition}

To see Definition~\ref{def: epsilon-net} and Definition~\ref{def: our epsilon-net} are equivalent, one can see the element $r$ of a range $r \in \mathcal{R}$ is a subset of $V$ which is an instance of the shrinking procedure. That is all the elements in $r$ shrink into a point in some graph $G_S$. Thus, an $\epsilon$-net should include one of the points in $r$, which leads to Definition~\ref{def: our epsilon-net}. The interpretation of $G_\epsilon$ is that it includes the majority of vertices (i.e., which is shrunk from $\epsilon n$ vertices) among all the $G_S$.

Notice that with an arbitrary cost function, one cannot have an $G_\epsilon$ or a $G_S$ with a small size. However, if the cost function has some structure, one can analyze the VC-dimension of the range space and utilize the theorem proved by Haussler-Welzl~\cite{haussler1986epsilon}, which states that any random sample set $S$ of $G$ with size 
\begin{equation}
\label{eq:Haussler-Welzl}
O(\frac{|VC|}{\epsilon} \log \frac{1}{\epsilon} + \frac{1}{\epsilon} \log \frac{1}{\delta})  
\end{equation}
is an $\epsilon$-net, with probaility more than $1-\delta$. In addition, we can use one sample set $S$ to fit all the cost functions. Thus, assume $\delta$ is a constant, the size of $S$ for the multi-cost function is $O(|VC|^*/\epsilon \log 1/\epsilon)$, where $|VC|^*$ is the largest VC-dimension among all the cost functions, which is highly scalable in respect to the graph size $n$ and the number of cost functions $k$.

\paragraph{Demonstration}
We demonstrate how to analyze the $VC$-dimension for a cost function $c$ with certain properties to generate the $\epsilon$-net. Assume function $c$ has a property that for every $(i,j) \in E$,  $\Delta(i,j) \leq c(i,j) \leq \kappa \Delta(i,j)$, where $\Delta(i,j)$ is the distance metric of the data space. Notice that this property is the same as discrete Lipschitz continuity~\cite{jiang2011free} except it also has the lower bound on $c(i,j)$. This property is commonly true when the cost function has a similar structure in the metric of the data space, L-norm class for instance. In real analysis, it can be interpreted that the cost of two points cannot be too high concerning the distance of the data space (e.g., the effort of increasing the income from 10,000 to 12,000 can not be too large) but also not too low if the distance of the data space is far (e.g., there should be non-negligible effort of increasing the income from 8,000 to 15,000). 

Now, observe that this property implies that the two points are $\kappa$-shrinkable if and only if they are within the distance of 1 in the data space. Thus, an instance of the shrinking procedure is the same as putting a ball with a radius $1/2$ in the data space and all the points in this ball can shrink into a point. We can see the range $\mathcal{R}$ is the balls in the data space and The $\epsilon$-net here is asking what is the smallest sample set so that any ball containing at least $\epsilon n$ points also contains one of the sampled points. Fortunately, the $VC$-dimension of shattering the points via balls is at most shattering the points via the hyperplanes, which has the VC-dimension as $d+1$. Via Haussler-Welzl's theorem in Equation~\ref{eq:Haussler-Welzl}, a random sampled set with size $O( (d+1)^*/\epsilon \log 1/\epsilon)$ is an $\epsilon$-net for function $c$.  


\section{EXPERIMENT}

To demonstrate our algorithm, we conducted experiments using two datasets. MNIST dataset~\cite{lecun2010mnist} and the Adult dataset~\cite{misc_adult_2}. The experiments have two parts. The first part is to show the ability of our approaches to find all the Pareto optimal recourse paths with multi-criteria in different scenarios. The second part shows the quality of the Pareto recourse paths under different numbers of random samples. This supports the argument that one can use random samples as the $\epsilon$-net with a certain number of samples.

\subsection{Scenarios}

\subsubsection{MNIST:} Our first scenario is to find a recourse path from one digit to another digit with the higher number where the number of each picture can never go lower(e.g. $3 \rightarrow 5 \rightarrow 7$ is allowed, but $3 \rightarrow 6 \rightarrow 5 \rightarrow 7 $ is not). We commence by extracting a subset from the original MNIST dataset~\cite{lecun2010mnist}. we define $cost_1$ as the absolute difference between the images. An actionable edge exists if and only if the number of the first image is lower than the second image, aiming for a recourse path where the number incrementally increases. $cost_2$ is the typical L2 distance between the images. The overall cost of $cost_1$ is the maximum value of $cost_1$ among all edges in the path, while the overall cost of $cost_2$ is the sum of $cost_2$ values among all edges.

\subsubsection{Adult:} The Adult dataset is widely employed in predictive modeling to determine whether an individual's income exceeds \$50,000 per year based on 14 features, such as age, education, occupation, hours worked per week, and others. The following outlines the specific steps and methodologies we employed in our experiment, our goal is to find the recourse path with an income of more than \$50,000.

We established an actionable dataset based on a set of predefined actionable criteria. During this process, some criteria were designated as immutable to reflect real-world conditions. For instance, in the Adult dataset, gender was considered immutable. From this actionable dataset, we randomly selected 256 instances that met the predefined actionable criteria for further analysis. A point with the lowest predicted probability by a simple one-layered MLP model was selected as the starting point.

To create the actionability graph, we utilized a KNN-graph (k=4) from the standardized dataset and applied our multi-cost shortest path algorithm to explore optimal changes in three key dimensions: age, education, and hours per week. We employed the non-standardized data to calculate the shortest paths based on these criteria, aiming to identify the most optimized change strategies within these three dimensions. To ensure a realistic application, we designed different cost functions for each criterion. The first cost is the percentage of the negative value of the Kernel Density Estimation (KDE) score, which is introduced in FACE~\cite{poyiadzi2020face}. This cost is also known as the negative log-likelihood (NLL), where a lower value signifies a higher density. The remaining costs are the L1 distances between the three key dimensions. Taking "age" as an example, when designing the cost function, it is notable that age cannot decrease during the entire recourse process.


\subsection{Multi-criteria paths}

\begin{figure*}[ht!]
     \centering
     \begin{subfigure}[b]{0.9\textwidth}
         \centering
         \includegraphics[width=0.6\linewidth]{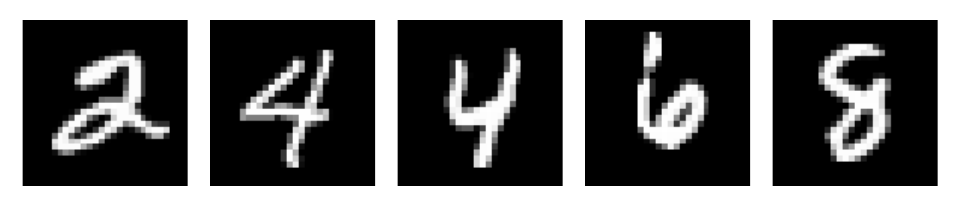}
         \caption{$Path$ with $cost_1$ \textbf{2}
         with $cost_2$ \textbf{31}.
         }
         \label{fig:y equals x}
     \end{subfigure}
     \hfill
     \begin{subfigure}[b]{0.9\textwidth}
         \centering
         \includegraphics[width=0.35\textwidth]{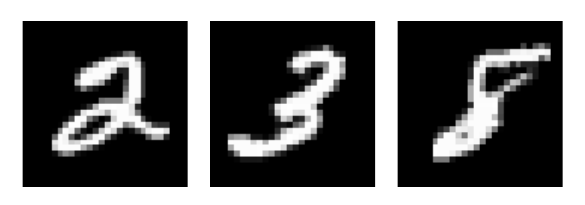}
         \caption{$Path$ with $cost_1$ \textbf{5}
         with $cost_2$ \textbf{15}.
         }
         \label{fig:three sin x}
    \end{subfigure}
     \hfill
     \begin{subfigure}[b]{0.9\textwidth}
         \centering
         \includegraphics[width=0.25\textwidth]{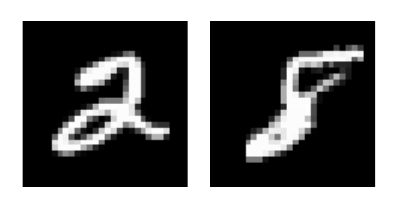}
         \caption{$Path$ with $cost_1$ \textbf{6}
         with $cost_2$ \textbf{8.7}.
         }
         \label{fig:five over x}
     \end{subfigure}
        \caption{The paths in MNIST under different criteria of $cost_1$ and $cost_2$.}
        \label{fig:three graphs}
        \vspace{0.3cm}
\end{figure*}

Figure~\ref{fig:three graphs} present all the Pareto optimal paths we found in MNIST with two cost functions $cost_1,cost_2$. The starting point is an arbitrary image labeled as 2 and we only sampled 256 images to construct the actionability graph. This is to simplify the graph for clear visualization. $cost_1$ represents the maximum number of digit transformations in the path. For example, in Path $(2,31)$, the maximum number of changes between images is at most 2. $cost_2$ is the sum of the L2-norm and the values have been standardized with the mean and variance among all the samples in the MNIST dataset. In this figure, one can see that by relaxing $cost_1$, the algorithm did find another path with lower $cost_2$, which demonstrates that Algorithm~\ref{algo:pareto-shortest} finds paths with different criteria.

\begin{figure}[ht]
    \centering
    \includegraphics[width=0.9\textwidth]{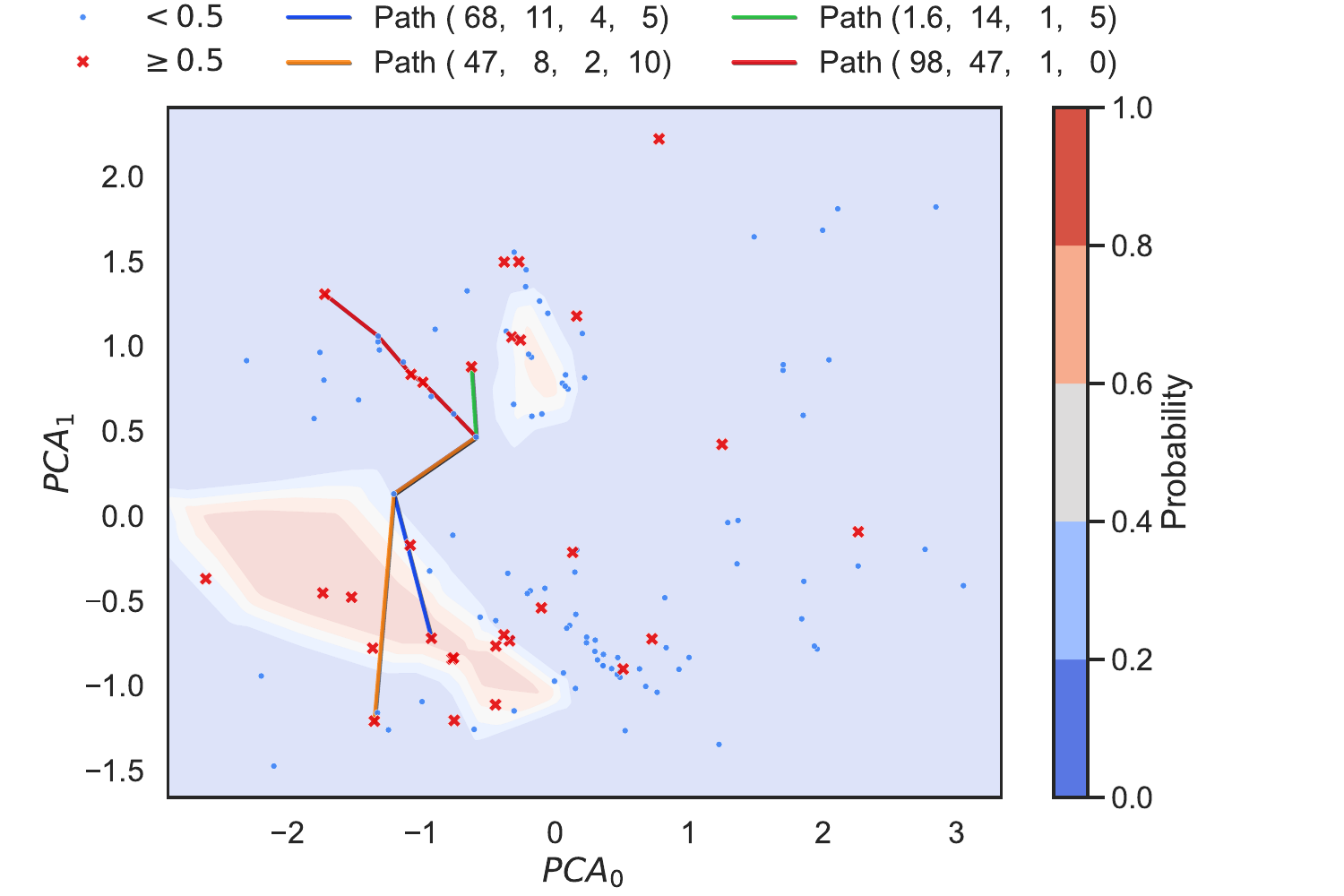}
    \caption{Multiple different paths from a starting data point to an end data point. It shows that the algorithm finds multi-criteria paths and can have a diverse path. The four paths show the trade-off between different criteria, which are the KDE, age, education-num, and hours-per-week.}
    \vspace{0.4cm}
    \label{fig:enter-label}
\end{figure}

Figure~\ref{fig:enter-label} visualized the results of the experiment in the adult dataset. The model is visualized with the probability output. Regions with more blue color are more likely to reject the object and regions with more red color are more likely to accept the object. However, since the original plot is in high-dimension we applied Principal Component Analysis (PCA) to reduce the dimension into two dimensions. The colors are just for indication and cannot be very accurate. The blue points represent data points with a success probability of less than 0.5, while the red points represent data points with a success probability greater than 0.75, which are also the endpoints. The numbers in the parentheses of the path represent age, education-num, hours-per-week, and KDE. The color bar and background colors indicate the estimated success probability, with redder hues signifying higher probabilities and bluer hues indicating lower probabilities.

These four paths, originating from the same starting point where the success probability is the worst, illustrate the diverse impacts caused by Pareto optimal solutions, and all four paths ultimately reach an endpoint where the success probability exceeds 0.75. The introduction of Kernel Density Estimation (KDE) into the algorithm demonstrates a better trade-off with other costs, enhancing the algorithm's ability to navigate through the solution space effectively.

\begin{figure}[ht!]
    \centering
    \includegraphics[width=0.8\textwidth]{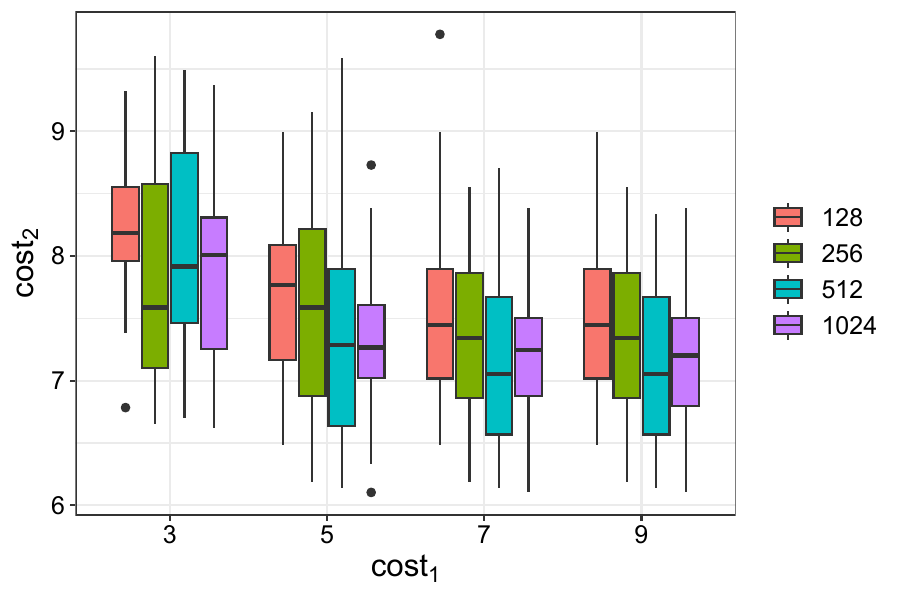}
    \caption{The trend among different sampling sizes shows that the quality of the Pareto paths becomes better and more stable when the sampling size gets larger. However, the improvement is not that significant when the size is large enough ($\geq 256$).}
    \label{fig:scalability}
    \vspace{0.2cm}
\end{figure}

\subsection{Scalability}
This experiment is designed to see the quality of the recourse paths when the graph is too large and we have to use $\epsilon$-net to run our algorithm. We use the MNIST dataset and randomly sample points with sizes $128,256,512,1024$ and run 32 trials. We then compare the Pareto optimal paths under different sampling sizes. Figure~\ref{fig:scalability} is the reported result. The x-axis is the criterion of $cost_1$ and the y-axis is the box-plot of $cost_2$ value among 32 trials. Generally, one can see that when the number of samples increased, the quality of the Pareto path was improved too. This is reasonable since the algorithm is more likely to find a better path for lower $cost_2$. However, the improvement is not that clear when the number of samples is large. For example, in the sampling size of 512 (blue box), the average value of $cost_2$ when $cost_1=7$ and $cost_1=9$ is smaller than the ones with a sampling size of 1024 (purple box). We believe that is because the influence of the sampling size on the quality is not that significant and the reported $cost_2$ value is fickle to the random sampling process. Responding to the discussion in Section~\ref{sec:scalability}, this supports the argument of the $\epsilon$-net which shows a random sampling can bound the quality when the size is large enough.

\section{CONCLUSION}
In conclusion, our algorithm proposes a novel idea that can be generalized to non-differentiable or discrete cost functions in the field of recourse. The algorithm is applied to multi-cost scenarios where users can combine their background knowledge in professional fields, set the most suitable cost function, and then finally get a more practical counterfactual. We show that our algorithm can find all the Pareto recourse plans optimally and can be scaled to a large graph with the utilization of $\epsilon$-net. We conduct experiments on two data sets separately. For the MNIST data set, we compared the performance of Pareto optimal paths under different sampling sizes and showed the result that transfers from one handwriting number into another and offers the whole process. For Adults, we show the diversity in each cost function, and those diverse cost functions finally lead to multi-Pareto optimal solutions.

\emph{Acknowledgement:} This work is supported by NSTC:113-2221-E-008-086.



\bibliographystyle{splncs04}
\bibliography{ref.bib}

\end{document}